\documentstyle{article}

\def\noheaderplainsetup{%

\topmargin=0pt \headheight=0pt \headsep=0pt  \oddsidemargin=0pt \evensidemargin=0pt  \textheight=8.9truein \textwidth=6.5truein}

\noheaderplainsetup

\begin{document}


\newcommand{\chess}{\mbox{\em Chess}}
\newcommand{\checkers}{\mbox{\em Checkers}}
\newcommand{\elzz}[2]{ \langle #1\rangle\hspace{-2pt} \downarrow\hspace{-2pt} #2} 
\newcommand{\elz}[1]{\mbox{$\parallel\hspace{-3pt} #1 \hspace{-3pt}\parallel$}} 
\newcommand{\propel}{\mbox{\bf CL1}}
\newcommand{\propell}{\mbox{\bf CL2}}
\newcommand{\propellc}{\mbox{\bf CL2}^\circ}
\newcommand{\propseqc}{\mbox{\bf CL9}^\circ}
\newcommand{\propseqcw}{\mbox{\bf CL9}^{\circ,\Phi}}
\newcommand{\propseq}{\mbox{\bf CL9}}
\newcommand{\propseqw}{\mbox{\bf CL9}^\Phi}
\newcommand{\predseq}{\mbox{\bf CL11}}
\newcommand{\propseqe}{\mbox{\bf CL10}}
\newcommand{\propseqec}{\mbox{\bf CL10}^\circ}
\newcommand{\propseqecc}{\overline{\mbox{\bf CL10}^\circ}}
\newcommand{\propseqecn}{\overline{\mbox{\bf CL10}^\circ}}
\newcommand{\emptyrun}{\langle\rangle} 
\newcommand{\oo}{\bot}            
\newcommand{\pp}{\top}            
\newcommand{\xx}{\wp}               
\newcommand{\legal}[2]{\mbox{\bf Lr}^{#1}_{#2}} 
\newcommand{\Legal}[1]{\mbox{\bf LR}^{#1}} 
\newcommand{\win}[2]{\mbox{\bf Wn}^{#1}_{#2}} 
\newcommand{\seq}[1]{\langle #1 \rangle}           


\newcommand{\intimpl}{\mbox{\hspace{2pt}$\circ$\hspace{-0.14cm} \raisebox{-0.043cm}{\Large --}\hspace{2pt}}}
\newcommand{\sintimpl}{\mbox{\hspace{2pt}\raisebox{0.033cm}{\tiny $ | \hspace{-4pt} >$}\hspace{-0.14cm} \raisebox{-0.039cm}{\large --}\hspace{2pt}}}
\newcommand{\ade}{\mbox{\Large $\sqcup$}}      
\newcommand{\ada}{\mbox{\Large $\sqcap$}}      
\newcommand{\sst}{\mbox{\raisebox{-0.07cm}{\scriptsize $-$}\hspace{-0.2cm}$\pst$}}
\newcommand{\scost}{\mbox{\raisebox{0.20cm}{\scriptsize $-$}\hspace{-0.2cm}$\pcost$}}
\newcommand{\sqc}{\mbox{\small \raisebox{0.0cm}{$\bigtriangleup$}}}
\newcommand{\sqci}{\mbox{\tiny \raisebox{0.0cm}{$\bigtriangleup$}}}
\newcommand{\sqd}{\mbox{\small \raisebox{0.06cm}{$\bigtriangledown$}}}
\newcommand{\sqdi}{\mbox{\tiny \raisebox{0.05cm}{$\bigtriangledown$}}}
\newcommand{\sqe}{\mbox{\large \raisebox{0.07cm}{$\bigtriangledown$}}}
\newcommand{\sqa}{\mbox{\large \raisebox{0.0cm}{$\bigtriangleup$}}}
\newcommand{\mld}{\vee}    
\newcommand{\mlc}{\wedge}  
\newcommand{\mle}{\mbox{\Large $\vee$}}    
\newcommand{\mla}{\mbox{\Large $\wedge$}}  
\newcommand{\add}{\sqcup}                      
\newcommand{\adc}{\sqcap}                      
\newcommand{\gneg}{\neg}                  
\newcommand{\rneg}{\neg}               
\newcommand{\pneg}{\neg}               
\newcommand{\mli}{\rightarrow}                     
\newcommand{\intf}{\$}               
\newcommand{\tlg}{\bot}               
\newcommand{\twg}{\top}               

\newcommand{\pst}{\mbox{\raisebox{-0.01cm}{\scriptsize $\wedge$}\hspace{-4pt}\raisebox{0.16cm}{\tiny $\mid$}\hspace{2pt}}}
\newcommand{\cla}{\mbox{\large $\forall$}}      
\newcommand{\cle}{\mbox{\large $\exists$}}        
\newcommand{\pintimpl}{\mbox{\hspace{2pt}\raisebox{0.033cm}{\tiny $>$}\hspace{-0.18cm} \raisebox{-0.043cm}{\large --}\hspace{2pt}}}
\newcommand{\pcost}{\mbox{\raisebox{0.12cm}{\scriptsize $\vee$}\hspace{-4pt}\raisebox{0.02cm}{\tiny $\mid$}\hspace{2pt}}}
\newcommand{\st}{\mbox{\raisebox{-0.05cm}{$\circ$}\hspace{-0.13cm}\raisebox{0.16cm}{\tiny $\mid$}\hspace{2pt}}}
\newcommand{\cost}{\mbox{\raisebox{0.12cm}{$\circ$}\hspace{-0.13cm}\raisebox{0.02cm}{\tiny $\mid$}\hspace{2pt}}}


\newtheorem{theoremm}{Theorem}[section]
\newtheorem{conjecturee}[theoremm]{Conjecture}
\newtheorem{exercisee}[theoremm]{Exercise}
\newtheorem{definitionn}[theoremm]{Definition}
\newtheorem{lemmaa}[theoremm]{Lemma}
\newtheorem{propositionn}[theoremm]{Proposition}
\newtheorem{conventionn}[theoremm]{Convention}
\newtheorem{examplee}[theoremm]{Example}
\newtheorem{remarkk}[theoremm]{Remark}
\newtheorem{factt}[theoremm]{Fact}

\newenvironment{conjecture}{\begin{conjecturee}}{\end{conjecturee}}
\newenvironment{definition}{\begin{definitionn} \em}{ \end{definitionn}}
\newenvironment{theorem}{\begin{theoremm}}{\end{theoremm}}
\newenvironment{lemma}{\begin{lemmaa}}{\end{lemmaa}}
\newenvironment{proposition}{\begin{propositionn} }{\end{propositionn}}
\newenvironment{convention}{\begin{conventionn} \em}{\end{conventionn}}
\newenvironment{remark}{\begin{remarkk} \em}{\end{remarkk}}
\newenvironment{proof}{ {\bf Proof.} }{\  $\Box$ \vspace{.1in} }
\newenvironment{example}{\begin{examplee} \em}{\end{examplee}}
\newenvironment{exercise}{\begin{exercisee} \em}{\end{exercisee}}
\newenvironment{fact}{\begin{factt} \em}{\end{factt}}

\title{ Computability-logic web: an alternative to deep learning}
\author{Keehang Kwon   \\  
\\  Department of Computing Sciences, DongA University, Korea.\\
 Email: khkwon@dau.ac.kr.}
\date{}
\maketitle
\begin{abstract} 
 {\em Computability logic} (CoL) is a powerful, mathematically rigorous computational model.
In this paper,  we show that CoL-web, a web extension to CoL,
naturally supports   web
programming  where database updates are involved. 
To be specific, we discuss an implementation of the AI ATM
 based on CoL ($\propseq$ to be exact).

More importantly, we  argue that CoL-web supports a general AI and, therefore, is a good alternative to neural nets and deep learning. We also discuss how to integrate neural nets into CoL-web.
 \end{abstract}


\section{Introduction}\label{sintr}

It is not dfficult to point out the weaknesses of neural nets and deep learning.
 Simply put, neural nets are too weak to support general AI.
 They receive inputs (numbers), perform  simple arithmetic operations  and produce outputs (numbers).
 Consequently, they  provide only primitive services such as object classifications.
 Although object classification has some interesting applications, the power of classification 
 is in fact not much
 compared to all the complex services a human can provide. Complex
 services -- making a coffee, withdrawing  money from ATM, etc -- are not well supported by neural nets.
 In addition, their classification services are not  perfect, as
 they are only approximate.

 A human can provide complex services to  others.
 The notion of services and how to complete them thus play a key role for an AI to imitate a human. 
 In other words, the 
right move towards general AI would be to find (a) a mathematical notion for services, and
 (b) how an AI  automatically generates a strategy for completing the service calls.

 Fortunately, Japaridze developed a theory for services/games involving complex ones.
{\em Computability logic} (CoL) \cite{Jap03}-\cite{Japfin}, is an
elegant theory of (multi-)agent services. In CoL, services are seen as games between a machine and its environment and logical operators stand for operations on games.
It understands interaction among agents in its most general --- game-based --- sense. 

In this paper, we discuss a web programming  model based on CoL and implement an AI ATM.
An AI ATM is different from a regular ATM  in that the former automatically generates a 
strategy for a service call, while the latter does not.

We assume the following in our model:

\begin{itemize}

\item Each agent corresponds  to a web site with a URL. An agent's  knowledgebase(KB) is described in its homepage.

\item  Agents are initially inactive. An  inactive agent  becomes activated  when another agent  invokes a query  for the former.

\item  Our model supports  the query/knowledge duality, also known as   querying knowledge.
  That is, knowledge of an agent can be obtained from another agent by invoking
  queries to the latter.
\end{itemize}

To make things simple, we choose \propseq -- a fragment of CoL -- as our target language.  \propseq\ includes
sequential operators: sequential disjunction ($\sqd$) and 
sequential conjunction ($\sqc$) operators. These operators model knowledgebase
updates.
Imagine an ATM that maintains balances on Kim. Balances change over time. Whenever Kim
presses the deposit button for \$1, the machine must be able to update the balance of the person.
This can be represented by

\[ balance(\$0)\sqc blance(\$1)\sqc \ldots\sqc. \]

In this paper, we present $\propseqw$
which is a  web-based implementation of \propseq. This implementation is
straightfoward and  its correctness is rather obvious.
What is interesting is that \propseq\ is a novel web
programming model with possible database updates.
It would
  provide a good starting point for future
 high-level web programming.

\section{Preliminaries}\label{s2}

In this section a  brief overview of $\propseq$ is given. 

There are two players: the machine $\pp$ and the environment $\oo$.

There are two sorts of atoms: {\em elementary} atoms $p$, $q$, \ldots to represent elementary games, and {\em general atoms} $P$, $Q$, \ldots to represent any, not-necessarily-elementary, games. 

\begin{description}
\item[Constant elementary games]  $\twg$ is always a true proposition, and $\tlg$ is always a false proposition.

\item[Negation]
 $\gneg$ is a role-switch operation: For example, $\gneg (0=1)$ is true,
while $(0=1)$ is false.

\item[Choice operations]
The choice operations model decision steps in the course of interaction, with disjunction $\add$ 
 meaning the machine's choice, and conjunction  $\adc$  
meaning choice by the environment. 

\item[Parallel operations]
 $A\mlc B$ means the parallel-and, while
 $A\mld B$ means the paralle-or. 
 In $A\mlc B$, $\pp$ is considered the winner if it wins in both $A$ and $B$, 
while in $A\mld B$ it is sufficient to win in one of $A$ and $B$.

\item[Reduction]
 $\mli$ is defined  by $\gneg A\mld B$.

\item[Sequential operations]
  $A\sqd B$ (resp. $A\sqc B$) is a game that starts and proceeds as a play of $A$; it will also end as an ordinary play of $A$ unless, at some point, 
$\pp$ (resp. $\oo$) decides --- by making a special {\em switch} move --- to abandon $A$ and switch to $B$.   $A\sqd B$ is quite similar 
 to the $if$-$then$-$else$ in imperative
languages. 

\end{description}

We reserve $\S$ as a special symbol for  {\bf switch moves}.
Thus, whenever $\oo$ wants to switch from a given component $A_i$ to $A_{i+1}$ in $A_0\sqc\ldots\sqc A_n$, it makes the move $\S$.  Note that
 $\pp$, too, needs to make switch moves in a $\sqc$-game to ``catch up" with $\oo$.
 The switches made by $\oo$ in a $\sqc$-game we call {\bf leading switches}, and the switches made by $\pp$ in a $\sqc$-game  we call {\bf catch-up switches}.

\section{Logic $\propseqw$}\label{intr}
In this section we review  
the propositional system $\propseq$ \cite{Japseq} and slightly extend it. 
Our presentation closely follows the one in \cite{Japseq}.
We assume that there are infinitely many nonlogical {\bf elementary atoms}, 
denoted by $p,q,r,s$ and
infinitely many nonlogical {\bf general atoms}, denoted by $P,Q,R,S$.

{\bf Formulas}, to which we refer as {\bf $\propseq$-formulas}, are built from atoms and operators in the standard way. 

\begin{definition}
  The class of $\propseq$-formulas 
is defined as the smallest set of expressions such that all atoms are in it and, if $F$ and $G$ are in it, then so are 
$\gneg F$, $F\mlc G$, $F \mld G$, $F \mli G$, $F\adc G$, $F \add G$, 
$F\sqc G$, $F \sqd G$. 
\end{definition}

Now we define $\propseqw$, a slight extension to $\propseq$ with environment parameters.
Let $F$ be a  $\propseq$-formula.
We introduce a new {\it env-annotated} formula $F^\omega$ which reads as `play $F$ against an agent $\omega$.
For an $\adc$-occurrence or an $\sqc$-occurrence $O$ in $F^\omega$, we say
$\omega$ is the {\it matching} environment of $O$.
For example, $(p \adc (q \adc r))^{w}$ is an  agent-annotated formula and
$w$ is the matching environment of both occurrences of $\adc$.
Similarly for $(p \sqc (q \sqc r))^{w}$.
We extend this definition to subformulas and formulas. For a subformula $F'$ of the above $F^\omega$,
we say that $\omega$ is the {\it matching} environment of both $F'$ and $F$.

In introducing environments to a formula $F$, one issue is
whether we allow `env-switching' formulas of the form
$(F[R^u])^w$. Here $F[R]$ represents a formula with some occurrence of a subformula $R$.
That is, the machine initially plays $F$ against agent $w$ and then switches
to play against another agent $u$ in the course of playing $F$.
For technical reasons,  we focus on non `env-switching' formulas.
This leads to the following definition: 

\begin{definition} 
  The class of $\propseqw$-formulas
is defined as the smallest set of expressions such that
(a) For any $\propseq$-formula $F$ and any agent $\omega$, $F^\omega$  are in it and, (b) if $H$ and $J$ are in it, then so are 
$\gneg H$, $H\mlc J$, $H\mld J$, $H\mli J$.
\end{definition}

\begin{definition}
\noindent Given a $\propseqw$-formula $J$,  the skeleton  of $J$ -- denoted by
$skeleton(J)$ -- is obtained by
replacing every occurrence $F^\omega$ by $F$.
\end{definition}
\noindent For example, $skeleton((p \adc (q \adc r))^{w}) = p \adc (q \adc r)$.

We borrow the following definitions  from \cite{Japseq}. They  apply both
to $\propseq$ and $\propseqw$.

An {\bf interpretation}\label{z6} for  $\propseq$ is a function that sends each nonlogical elementary atom  to an elementary game, and sends each general atom to any, not-necessarily-elementary, static game. This mapping extends to all formulas by letting it respect all logical operators as the corresponding game operations. That is, $\twg^*=\twg$, $(E\sqc F)^*=E^*\sqc F^*$, etc. When $F^*=A$, we say that {\bf $^*$ interprets $F$ as $A$}.

A formula $F$ is said to be {\bf valid} iff, for every interpretation $^*$, the game $F^*$ is computable. And $F$ is {\bf uniformly valid} iff there is an HPM $\cal H$, called a {\bf uniform solution} for $F$, such that $\cal H$ wins  $F^*$ for every interpretation $^*$. 

A {\bf sequential (sub)formula} is one of the form $F_0\sqc\ldots\sqc F_n$ or $F_0\sqd\ldots\sqd F_n$. We say that $F_0$ is the {\bf head} of such a (sub)formula, 
and $F_1,\ldots, F_n$ form its {\bf tail}. 

The {\bf capitalization} of a formula is the result of replacing in it every sequential subformula by its head.
 
 A formula is said to be  {\bf elementary} iff it is a formula of classical propositional logic.

An occurrence of a subformula in a formula is {\bf positive} iff it is not in the scope of $\gneg$. Otherwise it is {\bf negative}. 
 
A {\bf surface occurrence} is an occurrence that is not in the scope of a choice connective and not in the tail of any sequential subformula. 

The {\bf elementarization} of a $\propseq$-formula $F$ means the result of replacing in the capitalization of $F$ every surface occurrence of the form $G_1\adc\ldots\adc G_n$ by $\twg$, every surface occurrence of the form $G_1\add\ldots\add G_n$ by $\tlg$, and every positive surface occurrence of each general literal by $\tlg$. 

Finally,  a formula is said to be {\bf stable}\label{z14} iff its elementarization is a classical tautology; otherwise it is {\bf instable}.\vspace{10pt}

The proof system of $\propseqw$ is identical to that $\propseq$ in that
agent parameters play no roles.
$\propseqw$ consists of  the following four rules of inference.
\begin{definition}\label{defcl9}

\begin{description}
\item[Wait:]  $\vec{H}\mapsto F$, where $F$ is stable and $\vec{H}$ is the smallest set of formulas satisfying the following two conditions: 
\begin{enumerate} 
\item whenever $F$ has a surface occurrence of a subformula $G_1\adc\ldots\adc G_n$ whose matching environment is $\omega$, for each 
$i\in\{1,\ldots,n\}$, $\vec{H}$ contains the result of replacing that occurrence in $F$ by $G_i^\omega$;
\item whenever $F$ has a surface occurrence of a subformula $G_0\sqc G_1\sqc\ldots\sqc G_n$ whose matching environment is $\omega$, $\vec{H}$ contains the result of replacing that occurrence in $F$ by $(G_1\sqc\ldots\sqc G_n)^\omega$.
\end{enumerate}
\item[Choose:]  $H\mapsto F$, where $H$ is the result of replacing in $F$ a surface occurrence of a subformula $G_1\add\ldots\add G_n$ whose matching environment is $\omega$ by $G_i^\omega$ for some $i\in\{1,\ldots, n\}$.
\item[Switch:]  $H\mapsto F$, where $H$ is the result of replacing in $F$ a surface occurrence of a subformula $G_0\sqd G_1\sqd\ldots\sqd G_n$ whose matching environment is $\omega$ by 
$(G_1\sqd\ldots\sqd G_n)^\omega$.
\item[Match:] $H\mapsto F$, where $H$ is the result of replacing in $F$ two --- one positive and one negative ---
surface occurrences of some general atom by a nonlogical elementary atom that does not occur in $F$.\vspace{10pt}
\end{description}
\end{definition}

\begin{example}\label{jan11} The following is a $\propseq$-proof of
 $(b0\sqc b1\sqc b2)^u \mli (b0\sqc b1\sqc b2)^w$:\vspace{3pt}

$\begin{array}{ll}
1.\  b2^u \mli  b2^w   & \mbox{(from $\{\}$ by Wait)};\\
2.\  b2^u \mli (b1\sqc b2)^w & \mbox{(from 1 by Switch)};\\
3.\ ( b1\sqc b2)^u \mli (b1\sqc b2)^w & \mbox{(from 2 by Wait)};\\
4.\ ( b1\sqc b2)^u \mli (b0\sqc b1\sqc b2)^w & \mbox{(from 3 by Switch)};\\
5.\ (b0\sqc b1\sqc b2)^u \mli (b0\sqc b1\sqc b2)^w & \mbox{(from 4 by Wait)};\\
\end{array}$
\end{example}

\section{Logic $\propseqcw$}

To facilitate the execution procedure, following \cite{Japseq},
we modify $\propseqw$ to obtain
$\propseqcw$. 
Unlike $\propseqw$, this new language allows   hyperformulas which contain
the following. 

\begin{itemize}

\item Hybrid atom: each hybrid atom is a pair consisting of a general atom $P$, called its {\bf general component}, 
and a nonlogical elementary atom $q$, called its {\bf elementary component}. We denote such a pair by $P_q$. 
It  keeps track of the exact origin of each such elementary atom $q$.

\item Underlined sequential formula:
It is introduced for us not to forget the earlier components of sequential subformulas when Switch or Wait are applied.
We  now require that, in every sequential (sub)formula, one of the components be {\bf underlined}.  

\end{itemize}

The formulas of this  modified language we call {\bf hyperformulas}. 
We borrow the following definitions from \cite{Japseq}.

By the {\bf general dehybridization} of a hyperformula $F$ we mean the $\propseq$-formula that results from $F$ by replacing in the latter every hybrid atom by its general component, and  removing all underlines in sequential subformulas.

A {\bf surface occurrence} of a subexpression in a given hyperformula $F$
 means an occurrence that is not in the scope of a choice operator, such that, if the subexpression occurs within a component of a sequential subformula, that component is underlined or occurs earlier than  the underlined component.  

An {\bf active occurrence} is an occurrence such that, whenever it happens to be within a component of a sequential subformula, that component is underlined.

An {\bf abandoned occurrence} is an occurrence such that, whenever it happens to be within a component of a sequential subformula, that component is  to the left of the underlined component of the same subformula.

An {\bf elementary hyperformula} is one not containing choice and sequential operators, underlines, and general and hybrid atoms. 

The {\bf capitalization} of a hyperformula $F$ is defined
as the result of replacing in it every sequential subformula by its underlined component, after which all underlines are removed. 

The {\bf elementarization} \[\elz{F}\label{z25}\] of a hyperformula $F$ is 
the result of replacing, in the capitalization of $F$, every surface occurrence of the form $G_1\adc\ldots\adc G_n$ by $\twg$, every surface occurrence of the form $G_1\add\ldots\add G_n$
by  $\tlg$, every  positive surface occurrence of each general literal by $\tlg$, and every surface occurrence 
of each hybrid atom by the elementary component of that atom. 

A hyperformula $F$ is {\bf stable} 
iff its elementarization $\elz{F}$ is a classical tautology; otherwise it is {\bf instable}.

A hyperformula $F$ is said to be {\bf balanced} iff,  for every hybrid atom $P_q$ occurring in $F$, the following two conditions are satisfied: 
\begin{enumerate}
\item $F$ has exactly two occurrences of $P_q$,  one  positive and the other
 negative, and both occurrences are surface occurrences;
\item the elementary atom $q$ does not occur in $F$, nor is it the elementary component of any hybrid atom occurring in $F$ other than $P_q$.  
\end{enumerate}

An active occurrence of a hybrid atom (or the corresponding literal) in a balanced hyperformula is {\bf widowed} iff the other occurrence of the same hybrid atom is abandoned.

We extend $\propseqw$ to $\propseqcw$. The language of $\propseqc$\ 
allows any balanced hyperformulas, which we also refer to as {\bf $\propseqcw$-formulas}. 

\begin{definition}\label{nov23}
Logic $\propseqcw$ is given by the following rules for balanced hyperformulas (below simply referred to as ``(sub)formulas''): 
\begin{description}
\item[Wait$^\circ$:]  $\vec{H}\mapsto F$, where $F$ is stable and $\vec{H}$ is the smallest set of formulas satisfying the following two conditions: 
\begin{enumerate} 
\item whenever $F$ has an active surface occurrence of a subformula $G_1\adc\ldots\adc G_n$ whose matching environment is $\omega$ , for each 
$i\in\{1,\ldots,n\}$, $\vec{H}$ contains the result of replacing that occurrence in $F$ by $G_i^\omega$;
\item whenever $F$ has an active surface occurrence of a subformula $G_0\sqc \ldots\sqc \underline{G_m}\sqc G_{m+1}\sqc\ldots\sqc G_n$ whose matching environment is $\omega$, $\vec{H}$ contains the result of replacing that occurrence in $F$ by $(G_0\sqc \ldots\sqc G_m\sqc \underline{G_{m+1}}\sqc\ldots\sqc G_n)^\omega$.
\end{enumerate}
\item[Choose$^\circ$:]  $H\mapsto F$, where $H$ is the result of replacing in $F$ an active surface occurrence of a subformula $G_1\add\ldots\add G_n$ whose matching environment is $\omega$ by $G_i^\omega$ for some $i\in\{1,\ldots, n\}$.
\item[Switch$^\circ$:]  $H\mapsto F$, where $H$ is the result of replacing in $F$ an active surface occurrence of a subformula 
\(G_0\sqd\ldots\sqd \underline{G_m}\sqd G_{m+1}\sqd\ldots\sqd G_{n}\)  
whose matching environment is $\omega$ by \( (G_0\sqd\ldots\sqd {G_m}\sqd \underline{G_{m+1}}\sqd\ldots\sqd G_{n})^\omega.\)
\item[Match$^\circ$:] $H\mapsto F$, where $H$ has two --- a positive and a negative --- active surface occurrences of some hybrid atom $P_q$, and $F$ is the result of replacing in $H$ both occurrences by $P$.\vspace{10pt}  
\end{description}
\end{definition}

An effective procedure that converts any $\propseqw$-proof of any formula $G$ into a $\propseqcw$-proof of $G$ is given in \cite{Japseq}.

\section{Execution Phase}\label{s22tog}

The machine model of $\propseqcw$ is designed to process only one query/formula at one time.
In distributed systems, however, it is natural for an agent to receive/process
multiple queries from different users. For this reason, we introduce multiple queries to our machine.
To do this, we assume that an agent maintains two queues:
the {\it income queue} $QI$   for storing a sequence of
new incoming  queries of the form $(Q_1,\ldots,Q_n)$ and
the {\it temporarily solved queue} $QS$ for storing
a sequence of temporarily solved queries of the form
$(KB_1\mli Q_1,\ldots,KB_n\mli Q_n)$.
Here each $Q_i$  is a query and each $KB_i$ is a knowledgebase.
A query $Q$ with respect to some
knowledgebase is {\it temporarily solved} if $Q$ is solved but $\pp$ has a remaining switch move in $Q$.
Otherwise $Q$ is said to be {\it completely solved}.

As expected, processing  real-time multiple queries  causes some
complications.  
To be specific, we  process $QI$ of the form $(Q_1,\ldots,Q_m)$
and $QS$ of the form $(KB_1\mli Q'_1,\ldots,KB_n\mli Q'_n)$ in the following way:

\begin{enumerate}

 \item  First stage is to initialize a temporary variable $NewKB$ to $KB$,
     
\item The second stage is to follow the  $loop$ procedure: 

\end{enumerate}

procedure $loop$:   \\

\begin{itemize}
 
\item Case 1: $QI$ is not empty:

The machine tries to solve $Q_1$ by calling $Exec(NewKB\mli Q_1)$. \\

\begin{itemize}
  
\item  If it fails, then report a failure,
remove $Q_1$ from $QI$
and repeat $loop$.

\item Suppose it is a success and $NewKB$ and $Q_1$ evolve to $NewKB'$ and $Q'_1$ after solving this query. We consider two cases. \\
(a) If it is completely solved, then report a success,
remove $Q_1$ from $QI$, update $NewKB$ to $NewKB'$ and repeat $loop$. 
(b) If it is temporarily  solved,
 then report a success,
remove $Q_1$ from $QI$, insert $NewKB'\mli Q'_1$ to $QS$,
update $NewKB$ to $NewKB'$ and
repeat $loop$. 
\end{itemize}

\item Case 2. $QI$ is empty and $QS$ nonempty: The machine tries to solve the first 
query   $KB_1\mli Q'_1$ in $QS$. 

\begin{itemize}
\item  If $KB=NewKB$, it means nothing has changed since the last check.
  Hence the machine waits
  for any change such as the environment's new move. 
  
\item  Otherwise, the machine  tries to solve $Q'_1$ with respect to $NewKB$.
It thus removes the above  query  from $QS$, adds $Q'_1$ to $QI$,
and repeat $loop$. \\
\end{itemize}

\item Case 3. $QI$ is empty and $QS$ is empty:
wait for  new incoming service calls. 

\end{itemize}

Below we will introduce an algorithm that executes
a formula $J$. The algorithm is a minor variant
of the one in \cite{Japseq} and contains two stages: \\

{\bf Algorithm Exec(J)}: \%  $J$ is a $\propseqcw$-formula \\

\begin{enumerate}

 \item Fix an  interpretation $^*$. First stage is to initialize a temporary variable $E$ to $J$,
 a position variable $\Omega$ to an empty position $\langle\rangle$.
Activate all the resource agents specified in $J$ by invoking proper queries to them.
That is, for each negative occurrence of an annotated formula $F^\omega$ in $J$, 
    activate $\omega$ by querying $F^\mu$ to $\omega$.  Here $\mu$ is the current machine; 
    On the other hand, we assume that all the querying agents -- which appear positively in $J$ --
    are already active. 
        
\item The second stage is to play $J$ according to the following $mainloop$ procedure
(which is a minor variant of \cite{Japseq}): 

\end{enumerate}

procedure $loop(Tree)$: \%  $Tree$ is a proof tree of $J$ \\

If $E$ is derived  by Choose$^\circ$ from $H$, the machine
 makes the move $\alpha$ whose effect is choosing $G_{i}$ in the $G_1\add\ldots \add G_n$ subformula of $E$. 
So, after making move $\alpha$, the machine
call $loop$ on $\seq{\Omega}H^*$. 
 Let $\omega$ be the matching environment. Then inform $\omega$ of the move  $\alpha$.

If $E$ is derived  by Switch$^\circ$ from $H$, then 
the machine makes the move $\alpha$ whose effect is making a switch in the \(G_0\sqd\ldots\sqd \underline{G_m}\sqd G_{m+1}\sqd\ldots\sqd G_{n}\) subformula.  
 So, after making move $\alpha$, the machine calls $loop$ on
 $\seq{\Omega,\pp\alpha}H^*$.

If $E$ is derived by Match$^\circ$ from $H$ through replacing the two (active surface) occurrences of a hybrid atom $P_q$ in $H$ by $P$, 
then the machine finds within $\Omega$ and copies, in the positive occurrence of $P_q$, all of the moves made so far by the environment in the negative occurrence of $P_q$, and vice versa. This series of moves brings the game down 
to $\seq{\Omega'}E^*=\seq{\Omega'}H^*$, where $\Omega'$ is result of adding those moves to $\Omega$.  So, now the machine calls $loop$ on
 $\seq{\Omega'}H^*$.

Finally, suppose $E$ is derived by Wait$^\circ$. Our machine keeps
 granting permission (``waiting'').

{\em Case 1}. $\alpha$ is a move whose effect is moving in some abandoned subformula or a widowed hybrid literal of $E$.  In this case, the machine calls $loop$ on $\seq{\Omega,\oo\alpha}E^*$.

{\em Case 2}. $\alpha$ is a move whose effect is moving in some active surface occurrence of a general atom in $E$.  Again, in this case, the machine calls $loop$ on $\seq{\Omega,\oo\alpha}E^*$.

{\em Case 3}. $\alpha$ is a move whose effect is making a catch-up switch in some active surface occurrence of a $\sqd$-subformula. 
 The machine calls $loop$ on $\seq{\Omega,\oo\alpha}E^*$.

{\em Case 4}. $\alpha$ is a move whose effect is making a move $\gamma$ in some active surface occurrence of a non-widowed hybrid atom. Let $\beta$ be the move whose effect is making the same move $\gamma$ within the other active surface occurrence of the same hybrid atom.   In this case, the machine makes the move $\beta$ and calls $loop$ on $\seq{\Omega,\oo\alpha,\pp\beta}E^*$.

{\em Case 5}: 
 $\alpha$ is a move whose effect is a choice of the $i$th component in an active surface occurrence of a subformula $G_1\adc\ldots\adc G_n$. Then the machine
calls $loop$ on $\seq{\Omega}H^*$, where $H$ is the result of replacing the above subformula by $G_i$ in $E$.  

{\em Case 6}: $\alpha$ signifies a (leading) switch move within an active surface occurrence of a subformula \[G_0\sqc\ldots\sqc \underline{G_m}\sqc G_{m+1}\sqc \ldots\sqc G_n.\] Then the machine makes the same move $\alpha$ (signifying making a catch-up switch within the same subformula), and calls $exec$  on 
$\seq{\Omega,\oo\alpha,\pp\alpha} H^*$, where $H$ is the result of replacing the above subformula by \[G_0\sqc\ldots\sqc G_m\sqc \underline{G_{m+1}}\sqc \ldots\sqc G_n.\]

\section{Examples}\label{sec:modules}

As an example of web system, we will look at the ATM of some bank.
It is formulated with the user, an ATM, a database, and  a credit company.
We assume the following:

\begin{itemize}

\item There are two kinds of agents: super agents  and regular agents.
Super agents are prefixed with $\$$.  For example, $\$kim$ is a super agent.
While regular agents behave according to
 the $Exec$ procedure, super agents behave unpredictably.

\item For simplicity, we assume the bank has only one customer named Kim.
Further, the balance is restricted to one of the three amounts: \$0, \$1 or \$2.

\item The database maintains balance information on Kim. 

\item Both the credit company and the ATM 
request  balance checking to the database.
  
\item The ATM has a (\$1) deposit button. Whenever pressed, it adds $\$1$ to the
account.

\end{itemize}  
  
The above can be implemented as follows:

\newenvironment{exmple}{
 \begingroup \begin{tabbing} \hspace{2em}\= \hspace{3em}\= \hspace{3em}\=
\hspace{3em}\= \hspace{3em}\= \hspace{3em}\= \kill}{
 \end{tabbing}\endgroup}
\newenvironment{example2}{
 \begingroup \begin{tabbing} \hspace{8em}\= \hspace{2em}\= \hspace{2em}\=
\hspace{10em}\= \hspace{2em}\= \hspace{2em}\= \hspace{2em}\= \kill}{
 \end{tabbing}\endgroup}


\begin{exmple}
\> $agent\ credit$. \% credit company  \\
\> $ (b0 \sqc b1 \sqc b2)^{db}$. \% b0 means the balance is \$0, and so on.  \\
\end{exmple}

\% Here, we assume that ATM usage charge is zero, meaning deposit = balance.
\begin{exmple}
\> $agent\ db$. \% database \\
\> $ (d0 \sqc d1 \sqc d2)^{m}$. \% d0 means the accumulated deposit is \$0, and so on.  \\
\> $ d0 \mli b0$.   \\
\> $ d1 \mli b1$.    \\
\> $ d2 \mli b2$.   \\
\end{exmple}

\begin{exmple}
  \> $agent\ m$. \% ATM machine\\
\> $ (d0 \sqc d1 \sqc d2)^{\$kim}$. \%  request deposit checking to kim\\
\> $ (b0 \sqc b1 \sqc b2)^{db}$. \%  request balance checking  to DB \\
\end{exmple}

\begin{exmple}
  \> $agent\ \$kim$. \% $\$kim$ is a super agent.  \\
  \ 
  \> $ (b0 \sqc b1 \sqc b2)^m$. \%  request balance checking to ATM\\

\end{exmple}

Now let us consider  the agent $kim$ and the agent $credit$. They both want
to know the balance of Kim's account. The initial balance checking will return
$b0$, meaning zero dollars. Later, suppose Kim deposits \$1. In this case,
the balance information on $db$ will be updated to one dollar and,
subsequently, the  response to balance checking by ATM, kim and the
credit company will be updated
to $b1$, as desired.

\newcommand{\nagent}{$\eta$-agent}
\newcommand{\nagents}{$\eta$-agents}

\section{Adding neural networks}\label{sintr}

The integration of neural nets and symbolic AI is often beneficial.
There are several ambitious approaches such as DeepProblog\cite{Deep}
and these approaches
try to combine both worlds
{\it within} a single agent. Unfortunately, these approaches considerably
increase the complexity of the machine.

Fortunately, in the multi-agent setting, this integration can be achieved
in a rather simple way by introducing a new kind of agents called
\nagent (neural-net agents).

There are now three kinds of agents:

\begin{itemize}

\item  regular agents who perform deductive reasoning, and

\item \nagents\ who perform inductive reasoning.

    \item super agents who are able to create resources.

  \end{itemize}

An \nagent\  is an agent which is designed to
 implement  low-level perceptions (visual data, etc)  via trainig.
That is, its knowledgebase is in neural-net form for easy training.
In the sequel, \nagents\ are prefixed with $\eta$.

We assume the following:

(1) the input/output of a neural network is encapsulated in the form of an atomic predicate,

(2) the output of a 
neural network is deterministic. Thus, we do not consider probability
here, and

(3) For simplicity, neural nets are specified {\it in logical form}
($\propseqw$ to be
exact), instead of in functional form.

An \nagent\ with its knowledgebase $N$
proceeds in two modes:

\begin{itemize}

\item When there is a query $A$, it proceeds in deductive mode by
  processing $A$ using $N$ via $\propseqw$ deduction.

  \item When it is idle, it trains itself on sample data by updating $N$.

  \end{itemize}

As an example, we will look at the program which, given image of an animal,
identifies its habitat. \\

We assume the following: 

\begin{itemize}

\item The regular agent $a$ implements
  the predicate $habitat(j,h)$ where $j$ is an
   image of an animal and $h$ is the main habitat of the animal.
   We consider two kinds of animals: lion and tiger.
   We assume that $j$ belongs to $S= \{ i_1,\ldots,i_3 \}$ where
   $S$ is a set of  three images of animals.

\item The \nagent\ $\eta d$ implements
  the predicate $animal(j,n)$ where $j$
  is image of an animal and $n$ is the corresponding animal.

\end{itemize}  
  
The above can be implemented as follows:

\newenvironment{exam}{
 \begingroup \begin{tabbing} \hspace{2em}\= \hspace{3em}\= \hspace{3em}\=
\hspace{3em}\= \hspace{3em}\= \hspace{3em}\= \kill}{
 \end{tabbing}\endgroup}


\begin{exam}
\> $agent\ a$. \% animal habitats  \\
\> \% Below is a  query to $\eta d$.  \\
\> $ ((animal(i_1,lion)\add animal(i_1,tiger)) \adc$ \\
\> $ (animal(i_2,lion)\add animal(i_2,tiger)) \adc$ \\
\> $ (animal(i_3,lion)\add animal(i_3,tiger)))^{\eta d}$.\\
\> \% Rules that maps animals to  the corresponding habitats.  \\
\> $ animal(i_1,tiger) \mli habitat(i_1,india)$.    \\
\> $ animal(i_2,tiger) \mli habitat(i_2,india)$.    \\
\> $ animal(i_3,tiger) \mli habitat(i_3,india)$.    \\
\> $ animal(i_1,lion) \mli habitat(i_1,senegal)$.    \\
\> $ animal(i_2,lion) \mli habitat(i_2,senegal)$.    \\
\> $ animal(i_3,lion) \mli habitat(i_3,senegal)$.    \\
\end{exam}

\% Given image $j$, the agent $\eta d$ produces the corresponding animal
via deep learning. We do not show the details here. \\
\begin{exam}
\> $agent\ \eta d$. \% \nagent  \\
\> $ \vdots$.   
\end{exam}

When the agent $\eta d$ is idle, it  trains itself and  updates
its knowledgebase by adjusting weights. 
Now let us invoke   a query $habitat(i_3,india)\add habitat(i_3,senegal)$
to the agent $a$ where $i_3$ is the image of some animal. 
To solve this query, the agent $a$ invokes another  query shown above
to the agent $\eta d$.  Now $\eta d$ switches from
training mode to deduction mode to solve this query.
Let us assume that the response from $\eta d$
is $animal(i_3,lion)$. Using the rules related to animal's habitats,
the agent $a$ will return $habitat(i_3,senegal)$ to the user.
Note that our agents behave just like real-life agents.
For example, a doctor typically trains himself when he is idle.
If there is a
service request, then he switches from training mode to deduction mode.


\begin{thebibliography}{99}



\bibitem{Jap03} G. Japaridze. {\em Introduction to computability logic}. {\bf Annals of Pure and Applied Logic} 123 (2003), pp. 1-99.

\bibitem{Japtocl1} G. Japaridze. {\em Propositional computability logic I}. {\bf ACM Transactions on Computational Logic} 7 (2006), No.2, pp. 302-330.

\bibitem{Japtocl2} G. Japaridze. {\em Propositional computability logic II}. {\bf ACM Transactions on Computational Logic} 7 (2006), No.2, 
pp.  331-362.


\bibitem{Japfin} G. Japaridze. {\em In the beginning was game semantics}. In: {\bf Games: Unifying Logic, Language and Philosophy}. O. Majer,
A.-V. Pietarinen and T. Tulenheimo, eds. Springer Verlag, Berlin (to appear).  Preprint is available at http://arxiv.org/abs/cs.LO/0507045


\bibitem{Japseq} G. Japaridze. {\em Sequential operators in computability
  logic}. {\bf Information and Computation} 206, No.12 (2008), pp. 1443-1475.

  \bibitem{Deep} R. Manhaeve et al.
    {\em DeepProbLog: neural probabilistic logic programming},
    {\bf  NIPS'18: Proceedings of the 32nd International Conference on Neural Information Processing Systems}, 2018,
pp.3753--3763.

\end{thebibliography}
\end{document}